\pdfoutput=1

\documentclass[11pt]{article}

\usepackage{authblk}
\usepackage{acl}

\usepackage{times}
\usepackage{latexsym}

\usepackage[T1]{fontenc}

\usepackage[utf8]{inputenc}

\usepackage{microtype}
\usepackage{amsmath}
\usepackage{import}
\usepackage{fancyvrb}
\usepackage{xspace} 
\usepackage{caption}
\usepackage{subcaption}
\usepackage{tabularray}
\UseTblrLibrary{booktabs}

\newcommand{\ptb}{PTB\xspace}
\newcommand{\ptbplus}{PTB-modified\xspace}

\newcommand{\ftd}{FTD\xspace}
\newcommand{\clin}{CLIN\xspace}
\newcommand{\clinPOS}{CLINpos\xspace}

\newcommand{\rep}{REPETITION\xspace}
\newcommand{\res}{RESTART\xspace}

\newcommand{\parserNOAUGS}{parserSTD\xspace}
\newcommand{\parserALLAUGS}{parserAUG\xspace}

\newcommand{\pos}{POS\xspace}
\newcommand{\evalb}{evalb\xspace}

\makeatletter
\newcommand\email[2][]%
   {\newaffiltrue\let\AB@blk@and\AB@pand
      \if\relax#1\relax\def\AB@note{\AB@thenote}\else\def\AB@note{\relax}%
        \setcounter{Maxaffil}{0}\fi
      \begingroup
        \let\protect\@unexpandable@protect
        \def\thanks{\protect\thanks}\def\footnote{\protect\footnote}%
        \@temptokena=\expandafter{\AB@authors}%
        {\def\\{\protect\\\protect\Affilfont}\xdef\AB@temp{#2}}%
         \xdef\AB@authors{\the\@temptokena\AB@las\AB@au@str
         \protect\\[\affilsep]\protect\Affilfont\AB@temp}%
         \gdef\AB@las{}\gdef\AB@au@str{}%
        {\def\\{, \ignorespaces}\xdef\AB@temp{#2}}%
        \@temptokena=\expandafter{\AB@affillist}%
        \xdef\AB@affillist{\the\@temptokena \AB@affilsep
          \AB@affilnote{}\protect\Affilfont\AB@temp}%
      \endgroup
       \let\AB@affilsep\AB@affilsepx
}
\makeatother

\title{Improved \pos tagging for spontaneous, clinical speech using data augmentation}

\author[1]{Seth Kulick}
\author[1]{Neville Ryant}
\author[2]{David J. Irwin}
\author[2]{Naomi Nevler}
\author[1]{Sunghye Cho}
\affil[1]{Linguistic Data Consortium, University of Pennsylvania}
\email{\texttt{\{skulick,nryant,csunghye\}@ldc.upenn.edu}}
\affil[2]{Penn Frontotemporal Degeneration Center, University of Pennsylvania}
\email{\texttt{\{naomine,dirwin\}@pennmedicine.upenn.edu}}

\date{}

\begin{document}
\maketitle

\begin{abstract}
This paper addresses the problem of improving \pos tagging of transcripts of speech from clinical populations. In contrast to prior work on parsing and \pos tagging of transcribed speech, we do not make use of an in domain treebank for training. Instead, we train on an out of domain treebank of newswire using data augmentation techniques to make these structures resemble natural, spontaneous speech. We trained a parser with and without the augmented data and tested its performance using manually validated \pos tags in clinical speech produced by patients with various types of neurodegenerative conditions. 
\end{abstract}

\def\UrlBreaks{\do\/\do-\do.}

\section{Introduction}
\label{sec:intro}
This paper addresses the problem of improving \pos tagging of transcripts of speech from clinical populations.  While parsing and \pos tagging of spontaneous speech has been addressed in previous works \citep{charniak2001edit,lou2019neural}, the clinical transcripts we focus on are quite different from even typical conversational speech; e.g., the frequency of dysfluencies in patients with Alzheimer's and related dementias is substantially higher \citep{Boschi2017,Cho2022}.

Two approaches to the related problem of parsing conversational speech with dysfluencies have been taken in earlier work.  Work such as \citet{charniak2001edit} has focused on specialized components for identifying dysfluencies, which could then be removed for less-problematic parsing . More recently,  work such as \citet{lou2019neural,lou2020improving} has shown that dysfluency detection can be successfully integrated with modern parsing  by training on a treebank of speech with dysfluency annotation \citep{mitchell1999treebank}.  

Our work described here is closer to the second approach, in that we do not use a separate dysfluency identification component, but with some important differences. First, since there is no in-domain treebank for this clinical speech,  we explore the use of data augmentation on an out-of-domain treebank for parsing.  Second, our goal is accurate \pos annotation rather than full parsing or dysfluency detection; \pos tags can then be used to distinguish between different patient phenotypes \citep{Cho2021a}. Third, we do not assume gold sentence segmentation (cf. to \citet{lou2019neural}).

\section{Data}
\label{sec:data}

\subsection{Picture descriptions}
\label{sec:data:transcripts}
Our corpus (\clin) consists of verbatim transcripts of Cookie Theft picture descriptions \citep{Goodglass1983} that were produced by predominantly patients with various forms of dementia and a small number of their healthy caregivers at the Penn Frontotemporal Degeneration Center of the University of Pennsylvania. The recordings have been collected from early 2000s to date and have been manually transcribed by trained annotators using standardized transcription guidelines. The study was approved by the Institutional Review Board of the Hospital of the University of Pennsylvania.

{\small
\begin{Verbatim}
the= the mother pl- the cleaning uh d-
um # I forget what  you call it 
\end{Verbatim}
}

The transcripts include three types of markup to indicate dysfluencies, as shown in the example above. A \texttt{=} indicates that the word is a repetition, a \texttt{-} indicates that it a partial word, and a \texttt{\#} token indicates the end of a restart. However, these markups are inconsistently applied throughout the transcripts.

\subsection{\pos annotation}
\label{sec:data:gold}
In order to evaluate our parser's performance, we randomly sampled 377 utterances (10,875 tokens) from this data to serve as a test set (\clinPOS). Gold \pos tags were assigned to each token by a trained linguist, who manually corrected \pos output generated by the parser trained using the architecture and data augmentations described in Sections \ref{sec:augmentations} and \ref{sec:model}. Our tagset is identical to the Penn Treebank (\ptb) \cite{marcus-etal-1993-building} tag set with three differences: (1)  partial words are indicated by a \texttt{PT} tag; (2) repetitions are indicated by appending \texttt{\_RE} to the regular function tag; (3) an additional metadata flag (`\#') was used for the \texttt{\#} markup. E.g.:

{\small 
 \begin{Verbatim}[commandchars=&\{\}]
the=   the mother pl- the cleaning uh
&textcolor{red}{DT_RE  DT    NN   PT   DT  NN      UH}

d- um # I    forget what  you  call  it 
&textcolor{red}{PT UH # PRP   VBP    WPO  PRP   VBP  PRP} 
\end{Verbatim}
}

\begin{table}[h]
    \centering
    \begin{tblr}{l r r r}
        \toprule
                  &  \# sents &  \# tokens &  tokens/sent \\
        \midrule
        \clin     &    16,985 &    334,746 &  19.71 (23.31) \\ 
        \clinPOS  &       377 &     10,875 &  28.85 (34.46) \\
        \bottomrule
    \end{tblr}
    \caption{Sentence counts and token counts for the full clinical dataset (\textbf{\clin}) and manually \pos-tagged subset (\textbf{\clinPOS}). The last column depicts mean and standard deviation of sentence length in tokens.}
    \label{tab:data}
\end{table}
\section{Data Augmentations}
\label{sec:augmentations}
The parser (Section~\ref{sec:model}) is trained using the standard \ptb training, dev, and eval split.\footnote{Sections 02-21, 22, and 23, respectively.} However, we apply various transformations to bring the training data closer to what is seen in the transcripts. At each epoch of training, the training corpus is transformed with the following augmentations applied randomly. See Appendix~\ref{app:augmentations}  for the details of thresholds used for their application.

\subsection{Augmentations on a single tree}
There are four augmentations applied within one tree. These are as follows:

\ \\
\noindent (1) \textbf{\sc vbz}:  \texttt{(VBZ is)} is changed to  \texttt{(VBZ 's)}

\noindent (2) \textbf{\sc repetition}: For each constituent \texttt{XP}, we  insert a \texttt{EDITED\_REP} node as a left sister to \texttt{XP}, with a yield consisting of a selection of the first leaves in \texttt{XP}. We leave  \texttt{XP} as a flat subtree, with equal signs appended to some of the words.   We also insert commas in the \texttt{EDITED\_REP} yield.  

\noindent (3) \textbf{\sc partial}: We insert before each leaf a partial word version of the word, with the part-of-speech tag \texttt{PT} and a word made up of 2 or more characters from the word, with a hyphen appended.

\noindent (4) \textbf{\sc filler}  We  insert before each leaf a new leaf with tag \texttt{UH} and a filler word (\texttt{uh}, \texttt{um}, \texttt{eh}, or \texttt{mhm}).

Figure \ref{fig:aug1} shows an example of all four transformations applied to one tree.  Of course, not every tree will have every transformation applied to it.

\begin{figure}[t]
{\small 
\begin{Verbatim}[commandchars=&\{\}]
(S
  (NP-SBJ (DT The) (NN percentage)
          (NN change))
  (VP (VBZ is)
    (PP-PRD (IN since)
      (NP (NN year-end))))
  (. .))
\end{Verbatim}
}
\hrulefill
{\small 
\begin{Verbatim}[commandchars=&\{\}]
(S
&textcolor{red}{  (EDITED_REP (DT The=)}
&textcolor{red}{              (NN percentage=)}
&textcolor{red}{              (, ,))}
  (NP-SBJ
    (DT The)
&textcolor{red}{    (UH um)}
    (NN percentage)
&textcolor{red}{    (PT chan-)}
    (NN change))
  (VP
&textcolor{red}{    (UH uh)}
&textcolor{red}{    (VBZ 's)}
&textcolor{red}{    (EDITED_REP (IN since)}
&textcolor{red}{                (NN year-end=))}
    (PP-PRD (IN since)
            (NP (NN year-end)))))  
 (. .))
\end{Verbatim}
}
\caption{Example of the tree-internal transformations.  Top: a tree from the Penn Treebank, Bottom: a modifed version of the same tree, with the changes in red.}
\label{fig:aug1}
\end{figure}

\begin{figure}
{\small
\begin{Verbatim}
(S (NP-SBJ (DT The)
           (ADJP (RB closely) (VBN held))
           (NNP Hertz) (NNP Corp.))
   (VP (VBD had)
       (NP (NP (JJ annual) (NN revenue))
            ...)))
\end{Verbatim}
}
\hrulefill
{\small
\begin{Verbatim}
(S (NP-SBJ (NNP Hertz) (NNP Equipment))
   (VP (VBZ is (NP-PRD (DT a) ...))))
\end{Verbatim}
}
\hrulefill
{\small
\begin{Verbatim}[commandchars=&\{\}]
(S &textcolor{red}{(EDITED_RES}
 &textcolor{red}{    (NP-SBJ (DT The)}
&textcolor{red}{             (ADJP (RB closely)}
&textcolor{red}{                   (VBN held))} 
&textcolor{red}{             (NNP Hertz)}
&textcolor{red}{             (# #)))}
   (S (NP-SBJ (NNP Hertz) (NNP Equipment))
      (VP (VBZ is (NP-PRD (DT a) ...))))
\end{Verbatim}
}
\caption{Example of restart augmentation.  A prefix of the first tree is used to create an \texttt{EDITED\_RES} node, attached to the second tree, to for the third tree, with the inserted restart in red.}
\label{fig:restart}
\end{figure}

\subsection{Augmentations on two or three trees}

Two augmentations apply to more than one tree. 

\noindent (1) \textbf{\sc combine}  Two or three trees are combined by making them children of a new \texttt{S} node, with {\tt (CC and)} between each tree, along with lower-casing the the first word of the non-initial trees (unless it was a proper noun) and removing the final punctuation from the non-final trees.  

\noindent (2) \textbf{\sc restart} This augmentation also combines two or tree trees, but with one tree cut truncated and wrapped under an \texttt{EDITED\_RES} node, to simulate a restart.  A random number of words are kept (between 3 and 6 inclusive), and all leaves to the right are removed, with the tree adjusted accordingly.  A \texttt{(\# \#}) leaf is randomly inserted at the end of the \texttt{EDITED\_RES}.  Figure \ref{fig:restart} shows an example of a RESTART combining two trees.   With three trees, the middle tree would be truncated to form an \texttt{EDITED\_RES} between the other two trees.

\section{Model}
\label{sec:model}

We use the parser model of \citet{kitaev-etal-2019-multilingual}, which represents a constituency tree $T$ as a set of labeled spans $(i, j, l)$, where $i$ and $j$ are a span's beginning and ending positions and $l$ is its label. Each tree is assigned a score $s(T)$, which is decomposed as a sum of per-span scores assigned using a neural network that takes a sequence of embeddings as input, processes these embeddings using a transformer-based encoder \citep{vaswani2017attention}, and produces a span score from an MLP classifier \citep{stern-etal-2017-minimal}. The highest-scoring valid tree is then found using a variant of the CKY algorithm. \pos tags are recovered using a separate classifier operating on top of the encoder output, which is jointly optimized with the span classifier.  For more details, see \citet{kitaev-klein-2018-constituency}. 

Our implementation is based on version 0.2.0 of the Berkeley Neural Parser, using roberta-base \citep{liu2019roberta}  embeddings. Appendix~\ref{app:model} includes details about training and hyperparameters.

\section{Experiments and Results}
\label{sec:results}

\subsection{\pos results on GOLD \ftd}
\label{sec:results:pos}

\begin{table}[h]
    \centering
    \begin{tblr}{l r r r}
        \toprule
        &    &   \SetCell[c=2]{c} F1  && \\
                 \cmidrule[lr=-0.3]{3-4} 
tag   &   \# gold   &  \parserNOAUGS & \parserALLAUGS \\
        \midrule
NN    &       1,563   &    96.75  &    98.36 \\
DT    &       1,398   &    98.08  &    98.32 \\
IN    &        916   &    94.68  &    94.51 \\
\textbf{VBZ}   &        \textbf{882}   &    \textbf{98.97}  &    \textbf{99.54} \\
PRP   &        750   &    98.53  &    98.73 \\
VBG   &        608   &    99.42  &    99.51 \\
\textbf{UH}    &        \textbf{553}   &    \textbf{71.25}  &    \textbf{95.97} \\
RB    &        530   &    80.51  &    84.06 \\
CC    &        520   &    99.23  &    99.33 \\
NNS   &        349   &    98.71  &    99.00 \\
VB    &        315   &    96.74  &    98.10 \\
JJ    &        293   &    89.00  &    93.13 \\
VBP   &        254   &    97.44  &    98.82 \\
TO    &        234   &    99.57  &    99.79 \\
\textbf{PT}    &        \textbf{157}   &     \textbf{0.00}  &    \textbf{88.63} \\
\midrule
total  &      10,831   &    93.19  &    96.69 \\
              \bottomrule
    \end{tblr}
    \caption{Breakdown by most frequent \pos tags for the parser trained with no/all augmentations. For each \pos tag the number of gold tokens and F1 is presented. Note that the \textit{total} row includes all \pos tags.}
    \label{tab:pos:ftdc}
\end{table}

We parsed the sentences from the \clinPOS sample in two ways - using the parser model trained on the unaugmented training section (\parserNOAUGS), and the model trained with all data augmentations (\parserALLAUGS). The parsed sentences were evaluated by the accuracy of their assigned \pos tags on this sample. Overall and per-tag F1 (for the most common \pos tags) of the parser are presented in Table~\ref{tab:pos:ftdc}.\footnote{The total number of tokens/tags in Table \ref{tab:pos:ftdc} (10,831) is different from that in Table \ref{tab:data} (10,875). There are 44 \# tokens with a \# tag that indicate restart endings.}  (A complete table is in Appendix~\ref{app:results}).

While F1 increases across the board, there are three tags (bolded in the table) where the increase is particularly notable. (1) \texttt{PT} does not exist in the \ptb, and so it is not surprising that  \parserALLAUGS had such an increase. (2) \texttt{UH} also had a substantial increase, due to the different pause items put in by the \texttt{\sc pause} augmentation, and (3)  \texttt{VBZ}, while already high with the \parserNOAUGS, further improved.

\subsection{Parser evaluation on augmented data}
\label{sec:results:ptbplus}
We are limited in the evaluation on the transcript data due to the lack of gold trees. To further evaluate the parsing with augmented training data, we created the \ptbplus versions of the  standard dev and test  sections, running 10 iterations of the augmentations to ensure a well-balanced selection of the various modifications. 
We parsed these \ptbplus dev and eval sections using both the \parserNOAUGS and \parserALLAUGS versions of the parser.  

The results are shown in the \ptbplus columns of Table ~\ref{tab:evalb:ptb}. The highlighted numbers show the  huge increase in accuracy going from \textbf{\parserNOAUGS} to \textbf{\parserALLAUGS}.  

\begin{table}[tb]
    \centering
    \begin{tblr}{l r r r r}
        \toprule
           & \SetCell[c=2]{c} \ptb &&  \SetCell[c=2]{c} \ptbplus && \\
             \cmidrule[lr=-0.3]{2-3} \cmidrule[lr=-0.3]{4-5} 
        system         &  \evalb  &  \pos    &  \evalb  &  \pos \\
        \midrule
        \SetCell[c=5]{c}\textit{development set} \\
        \parserNOAUGS       &  95.04 &  97.42 &  53.70 &  72.85 \\
        \parserALLAUGS       &  94.88 &  97.43 &  \textbf{95.18} &  \textbf{97.85} \\
        \SetCell[c=5]{c}\textit{evaluation set} \\
        \parserNOAUGS       &  95.40 &  97.85 &  53.89 &  73.10  \\
        \parserALLAUGS      &  95.37 &  97.70 &  \textbf{95.71} &  \textbf{98.06}  \\
        \bottomrule
    \end{tblr}
    \caption{Parser (\evalb F1) and \pos (accuracy) scores for the dev/eval sets of standard \ptb (\textbf{\ptb}) and the version of \ptb with data augmentations (\textbf{\ptbplus}).}
    \label{tab:evalb:ptb}
\end{table}

\subsection{Parser evaluation on standard data}
\label{sec:results:ptb}

However, we were also concerned with the impact of the augmentations on parsing more standard, non-clinical data.  To test these, we also parsed and evaluated the two parser models on the standard dev and test sections, with the results shown in the PTB columns in Table \ref{tab:evalb:ptb}. As can be seen, there is very little effect on the evalb or pos accuracy (e.g. going from 95.40\% to 95.37\% on the eval section). 

Therefore, the parser trained with the augmentations can parse the augmented versions of the dev and eval sections while having little decrease in accuracy on the standard non-augmented versions.

\subsection{Dysfluencies}
\label{sec:dysfluency}
While the focus of this work has been on increasing \pos accuracy on \clin with augmentations, we can also score how well these augmentations do at identifying the two types of dysfluences that can span multiple words - repeats and restarts.  The first row in  table \ref{tab:dysfluency:overall} shows the accuracy of \parserALLAUGS at capturing the repeats.  We scored this by post-processing the parser output to add \texttt{\_RE} to all tags in the yield of a \texttt{EDITED\_REP}, and then scoring which tags in the parser output and \clinPOS had \texttt{\_RE}. While the score is low  (56.54\%), we attribute to the inconsistency of the transcription in \clin.  While repeated words should have \texttt{=} appended to the text, this is the case for only 7\% of the \texttt{\_RE} words. 
 
 We also evaluated the parser's recovery of restarts by scoring intersections between the span of each \texttt{EDITED\_RES} nonterminal and our own annotation indicating the start of each restart. As restart detection F1 was only 16.07\%, we also trained a version of the parser in which the \texttt{\sc RESTART} augmentation was omitted for comparison. Interestingly, removing this augmentation caused a 4\% absolute drop in \rep detection F1. 

\begin{table}[t]
    \centering
    \begin{tblr}{l r r r}
        \toprule
        dysf.      &  restarts? &  \# gold &   f1  \\
        \midrule
        \rep       &          yes &   226     &  56.54 \\
                   &          no  &   226     &  52.59 \\
        \res       &          yes &    152    &  16.07 \\
        \bottomrule
    \end{tblr}
    \caption{F1 for repetition and restart dysfluency detection from parser output. Performance is reported both for \parserALLAUGS and for a variant in which the RESTART augmentation is omitted during training.}
    \label{tab:dysfluency:overall}
\end{table}

\section{Conclusion}
 \label{sec:conclusion}
 We have shown that data augmentation techniques can be used as a step toward adapting out-of-domain training material for parsing and \pos tagging of clinical speech, obtaining improved \pos tags across-the-board. However, it remains unclear as to how the performance of this data augmentation method compares to training on in-domain materials (when available). While no gold constituency parse trees exist for the clinical datatset described in Section~\ref{sec:data}, we do have gold trees for several hundred thousand tokens of Switchboard conversational telephone speech \citep{mitchell1999treebank}. One natural follow up is to directly compare our approach and that of training with in-domain trees \citep{lou2019neural} in terms of  \pos tagging, \evalb, and disfluency recovery on that corpus. Additionally, we intend to explore combining the two techniques to see if they are complementary and could be leveraged to further improve performance, both for Switchboard and for the clinical data described in this paper.
 
 We will also improve the data augmentation methods, both by moving away from hand-engineered frequencies, and by improving the \texttt{\sc restart} augmentation by bringing in external sources, instead of relying on an existing treebank.  For example, chatGPT could be used generate sentences from which we could then truncate for a restart text.

\section*{Acknowledgments}
This work was supported by the Department of Defense (W81XWH-20-1-0531), the National Institutes of Health (AG073510-01, P01-AG066597), and the Alzheimer's Association (AARF-21-851126). We would also like to acknowledge the contributions and support of the late Murray Grossman.

\bibliography{refs}

\appendix

\section{Data augmentation}
\label{app:augmentations}

\paragraph{\sc partial} For each token there is a 20\% chance of the partial word transformation applying. When this transformation applies, it takes into account the length of the word in characters:
    \begin{itemize}
        \item \textbf{n=3}: If the word has three characters, the first two
will be used for the partial word.
        \item \textbf{n=4}: If the word has four characters, it will make a partial word with 2 or 3 characters with equal probability.  
        \item \textbf{n>4}: If the word has more than 4 characters, 30\% of the time it will use 2 characters, 30\% of the time it 
will use 3 characters, and 40\% it will use 4 characters. 
    \end{itemize}  

\paragraph{\sc filler} There is a 10\% chance of inserting a filled pause after every word.

\paragraph{\sc repetition} For each constituent, there is a 30\% chance of the repetition transformation applying. When this transformation applies, a repetition of length 1, 2, or 3 words (with equal probability) is introduced as described in Section~\ref{sec:augmentations}.

Repeated words inserted by this transformation optionally are annotated with the ``='' suffix used to indicate repetitions in the \clin transcripts. This marker is add 70\% of the time to account for unreliability in usage by transcribers.

To account for variability among transcribers in use of punctuation, commas are optionally inserted between the words in an \texttt{EDITED\_REP} (20\% of the time). 

\paragraph{\sc vbz} A \texttt{(VBZ is)} is changed to \texttt{(VBZ 's)} 50\% of the time.

\paragraph{\sc combine, restart} As described in Section~\ref{sec:augmentations},  the \texttt{\sc combine} and \texttt{\sc restart} augmentations combine either 2 or 3 trees.  For each tree in the training section a decision is made to either not combine it with another tree, or to combine it with one or two of the trees after it in sequence as a \texttt{\sc combine} or \texttt{\sc restart}.

The percentages used are:
\begin{itemize}
    \item no combination - 65\%
    \item \texttt{\sc combine} with two trees - 20\%
    \item \texttt{\sc combine} with three trees - 10\%
    \item \texttt{\sc restart} with two trees - 2\%
    \item \texttt{\sc restart} with three trees - 3\%
\end{itemize}

For both versions of {\sc restart},  a \texttt{\#} is added at the end of the \texttt{EDITED\_RES} 50\% of the time to account for annotator variability.

\section{Model}
\label{app:model}
\begin{table}[t]
\footnotesize
    \centering
    \begin{tabular}{|l|r|} \hline
        hyperparameter & value \\ \hline
attention\_dropout & 0.2    \\ \hline
batch\_size & 32    \\ \hline
char\_lstm\_input\_dropout & 0.2    \\ \hline
checks\_per\_epoch & 4    \\ \hline
clip\_grad\_norm & 0.0    \\ \hline
d\_char\_emb & 64    \\ \hline
d\_ff & 2048    \\ \hline
d\_kv & 64    \\ \hline
d\_label\_hidden & 256    \\ \hline
d\_model & 1,024    \\ \hline
d\_tag\_hidden & 256    \\ \hline
elmo\_dropout & 0.5    \\ \hline
encoder\_max\_len & 512    \\ \hline
force\_root\_constituent & 'auto'    \\ \hline
learning\_rate & 5e-05    \\ \hline
learning\_rate\_warmup\_steps & 160    \\ \hline
max\_consecutive\_decays & 3    \\ \hline
max\_len\_dev & 0    \\ \hline
max\_len\_train & 0    \\ \hline
morpho\_emb\_dropout & 0.2    \\ \hline
num\_heads & 8    \\ \hline
num\_layers & 8    \\ \hline
predict\_tags & True    \\ \hline
relu\_dropout & 0.1    \\ \hline
residual\_dropout & 0.2    \\ \hline
step\_decay\_factor & 0.5    \\ \hline
step\_decay\_patience & 5    \\ \hline
tag\_loss\_scale & 5.0    \\ \hline
max\_epochs & 50    \\ \hline
\end{tabular}
    \caption{Hyperparameters used with the Berkeley Neural Parser.}
    \label{tab:hyperparameters}
\end{table}

Table \ref{tab:hyperparameters} shows the hyperparameter settings used in the Berkeley Neural Parser (all default). We added a parameter {\tt max\_epochs} for the maximum number of epochs, setting it to 50 for the cross-validation training reported.

\section{Results}
\label{app:results}

\subsection{Full \pos tagging scores}
Table \ref{app:pos:ftdc} shows a complete version of the information in Table \ref{tab:pos:ftdc}, with the scores for the \pos tags when parsed without and with the data augmentations.   Note that the tags with 0 gold counts at the bottom of the table indicate tags that are not in \clinPOS.  The occurred for several tags with the \parserNOAUGS parse since it mistakenly assigned such tags, in particular to the pauses.

\begin{table*}[t]
    \centering
    \SetTblrInner{rowsep=0pt}
    \begin{tblr}{l r r r|[dashed] l r r r}
        \toprule
        &    &   \SetCell[c=2]{c} F1  && &    &   \SetCell[c=2]{c} F1  && \\
                 \cmidrule[lr=-0.3]{3-4} \cmidrule[lr=-0.3]{7-8} 
        tag   &    \# gold   &  \parserNOAUGS & \parserALLAUGS & tag   &   \# gold   &  \parserNOAUGS & \parserALLAUGS \\
        \midrule
        NN    &       1563   &    96.75  &    98.36  &  WDT   &         59   &    89.08  &    89.47  \\
        DT    &       1398   &    98.08  &    98.32  &  WP    &         48   &    96.77  &    97.87  \\
        IN    &        916   &    94.68  &    94.51  &  WRB   &         13   &    86.67  &   100.00  \\
        VBZ   &        882   &    98.97  &    99.54  &  JJR   &         11   &   100.00  &   100.00  \\
        PRP   &        750   &    98.53  &    98.73  &  PDT   &          9   &    77.78  &    71.43  \\
        VBG   &        608   &    99.42  &    99.51  &  POS   &          9   &    62.07  &    84.21  \\
        UH    &        553   &    71.25  &    95.97  &  NNP   &          8   &    27.59  &    37.21  \\
        RB    &        530   &    80.51  &    84.06  &  RBR   &          7   &   100.00  &   100.00  \\
        CC    &        520   &    99.23  &    99.33  &  X     &          4   &     0.00  &     0.00  \\
        ,     &        409   &    99.39  &    99.39  &  XX    &          2   &     0.00  &     0.00  \\
        .     &        356   &    98.72  &    98.86  &  RPR   &          1   &     0.00  &     0.00  \\
        NNS   &        349   &    98.71  &    99.00  &  RBS   &          1   &   100.00  &     0.00  \\
        VB    &        315   &    96.74  &    98.10  &  notag &          1   &     0.00  &     0.00  \\
        JJ    &        293   &    89.00  &    93.13  &  JJS   &          1   &   100.00  &    66.67  \\
        VBP   &        254   &    97.44  &    98.82  &  :     &          0   &    0.00   &     -  \\
        TO    &        234   &    99.57  &    99.79  &  FW    &          0   &    0.00   &     -      \\
        PT    &        157   &     0.00  &    88.63  &  -LRB- &          0   &    0.00   &     -      \\
        PRP\$  &        123   &    95.97  &    96.75  &  ``    &          0   &   0.00   &    -       \\
        EX    &         97   &    97.38  &    96.41  &  ''    &          0   &    0.00   &     -      \\
        CD    &         89   &    95.03  &    96.13  &  SYM   &          0   &    0.00   &     0.00  \\
        MD    &         70   &    97.14  &    99.28  &  -RRB- &          0   &    0.00   &     -      \\
        VBD   &         66   &    93.43  &    94.74  &  NNPS  &          0   &    0.00   &     0.00  \\
        RP    &         65   &    67.80  &    64.41  &  LS    &          0   &    0.00   &     -      \\
        VBN   &         60   &    86.96  &    88.14  &  \#     &          0   &   0.00    &    -       \\
        \midrule
        \textbf{total}  &      \textbf{10,831}   &    \textbf{93.19}  &    \textbf{96.69} \\
        \bottomrule
    \end{tblr}
    \caption{A complete version of Table \ref{tab:pos:ftdc}, with a breakdown by most frequent \pos tags for the version of the parser trained without any data augmentation (\textbf{\parserNOAUGS}) and the parser trained using all augmentations (\textbf{\parserALLAUGS}). For each \pos tag the number of gold tokens and F1 is presented. Note that the \textit{total} row includes all \pos tags.}
    \label{app:pos:ftdc}
\end{table*}

\subsection{Repetition detection and dysfluency length}

We further evaluated the \rep recovery by breaking the 56.64 score down by length of repetition, shown in Table \ref{tab:dysfluency:re_detailed}.  Unsurprisingly, it did better the shorter the repetition was, but was still fairly stable with the repetitions of length one or two words, which comprise 78.3\% of the repetitions.

\begin{table}[t]
    \centering
    \begin{tblr}{l r r r r}
        \toprule
        length  &  \# gold  &  precision &  recall &  f1 \\
        \midrule
        1          &       121 &      62.30 &   62.81 &  62.55\\
        2          &        56 &      51.35 &   67.86 &  58.46 \\
        3          &        15 &      28.57 &   80.00 &  42.11\\
        $\geq$4    &        34 &      80.00 &   23.53 &  36.36 \\
        \midrule
        total      &       226 &      54.03 &   59.29 &  56.54\\
        \bottomrule
    \end{tblr}
    \caption{\rep detection (F1) for \parserALLAUGS as a function of dysfluency length in tokens.}
    \label{tab:dysfluency:re_detailed}
\end{table}

\end{document}